\documentclass{article}
\usepackage{spconf,amsmath,graphicx}

\usepackage{float}
\usepackage[utf8]{inputenc}
\usepackage{makeidx}
\usepackage{url}
\usepackage{epstopdf}
\usepackage{doc}
\usepackage{enumerate}
\usepackage{bbm}
\usepackage{multicol, blindtext}
\usepackage{ctable}
\usepackage{textcomp}	
\usepackage{multirow}

\usepackage[singlelinecheck=false]{caption}
\usepackage[]{footmisc}

\title{Convolutional Neural Networks with Layer Reuse}
\name{Okan K\"op\"ukl\"u, Maryam Babaee, Stefan H\"ormann, Gerhard Rigoll}
\address{Technical University of Munich \\
Institute for Human-Machine Communication \\
\small{\{okan.kopuklu, maryam.babaee, s.hoermann, rigoll\}@tum.de}}
\begin{document}
%
\maketitle
\begin{abstract}
A convolutional layer in a Convolutional Neural Network (CNN) consists of many filters which apply convolution operation to the input, capture some special patterns and pass the result to the next layer. If the same patterns also occur at the deeper layers of the network, why wouldn't the same convolutional filters be used also in those layers? In this paper, we propose a CNN architecture, Layer Reuse Network (LruNet), where the convolutional layers are used repeatedly without the need of introducing new layers to get a better performance. This approach introduces several advantages: (i) Considerable amount of parameters are saved since we are reusing the layers instead of introducing new layers, (ii) the Memory Access Cost (MAC) can be reduced since reused layer parameters can be fetched only once, (iii) the number of nonlinearities increases with layer reuse, and (iv) reused layers get gradient updates from multiple parts of the network. The proposed approach is evaluated on CIFAR-10, CIFAR-100 and Fashion-MNIST datasets for image classification task, and layer reuse improves the performance by 5.14\%, 5.85\% and 2.29\%, respectively. The source code and pretrained models are publicly available \footnote{https://github.com/okankop/CNN-layer-reuse}.
\end{abstract}


\section{Introduction}

\begin{figure}[t!]
	\centering
	\includegraphics[width=0.45\textwidth]{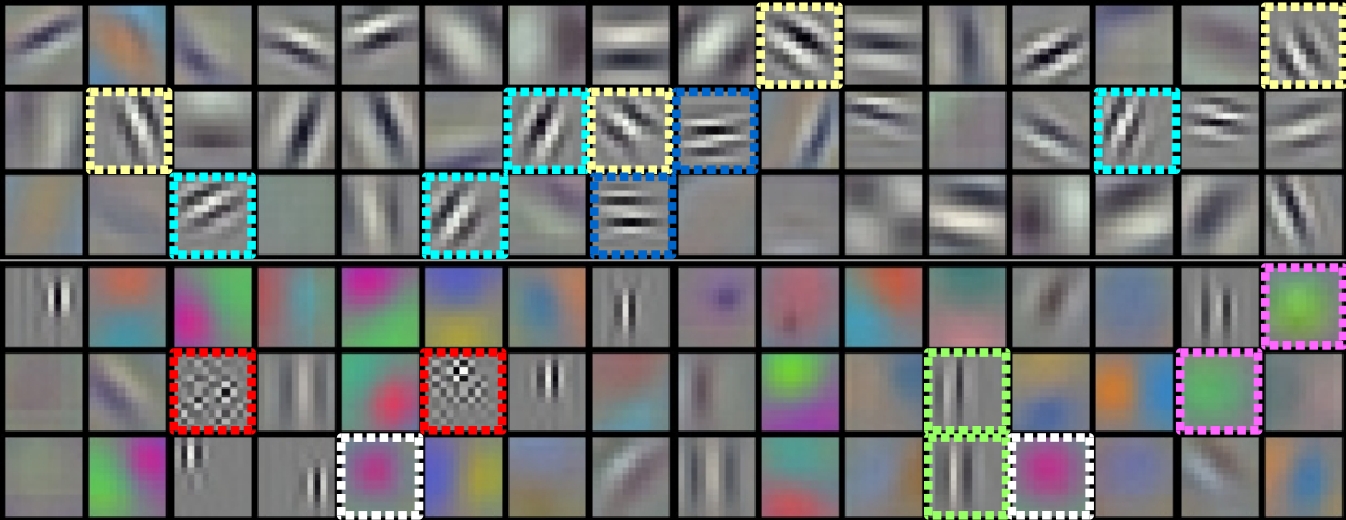}
	\caption{Learned convolutional kernels in the first convolutional layer of AlexNet. Some of the convolutional kernels are very similar, which are highlighted with the same color. Better view in color.}
	\label{fig:kernels}
\end{figure}

The conventional way of designing a Convolutional Neural Network (CNN) is to put convolutional, pooling and batch normalization layers one after another with some nonlinearity functions in between. With the invention of residual connections \cite{he2016deep}, deeper CNN architectures started to be trained without overfitting and they achieved better results compared to shallow ones. Consequently, the primary trend for solving major visual recognition tasks has become building deeper and larger CNNs \cite{krizhevsky2012imagenet, he2016deep, szegedy2015going}. The best performing CNNs usually have hundreds of layers and millions of parameters.  However, deeper architectures also mean more parameters which makes the designed architecture not appropriate for embedded and mobile devices. But the question always comes to mind: "Do we really need all these layers and parameters to achieve a better performance?". Fig. \ref{fig:kernels} shows the learned convolutional kernels in the first convolutional layer of AlexNet \cite{krizhevsky2012imagenet}. It is obvious to notice that some of the kernels are very similar hence redundant. Therefore, instead of introducing all these similar kernels separately, they can simply be reused.

 \begin{figure}[t!]
	\centering
	\includegraphics[width=0.4\textwidth]{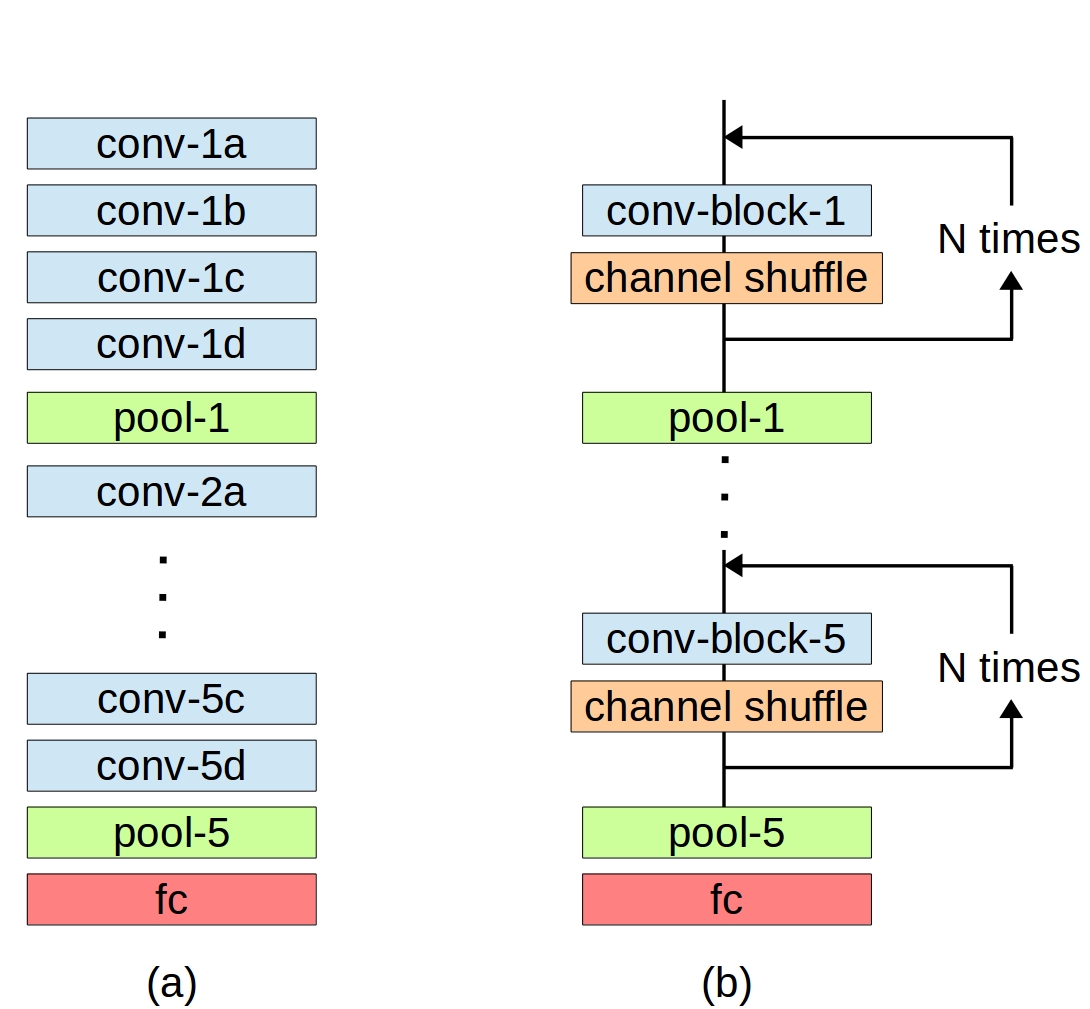}
	\caption{(a) Conventional design of CNNs, (b) CNN design with layer reuse. Instead of stacking convolutional layers and feeding one layer's output as input to another layer, we feed the output of a convolutional block as input to itself for \textit{N}-times before passing it to the next block.}
	\label{fig:lru_arch}
\end{figure}


In this paper, we propose a network architecture, Layer Reuse Network (LruNet), where we have reused some convolutional layers repeatedly. Instead of stacking convolutional layers one after another, as in Fig. \ref{fig:lru_arch} (a), we feed the output of the convolutional blocks to itself for a given \textit{N} times before passing the output to the next layer, as in Fig. \ref{fig:lru_arch} (b). While doing this, we apply channel shuffling in order to ensure feeding the outputs of convolution filters as inputs to other filters in the same block. Layer reuse (LRU) brings several advantages to the system: (i) The number of parameters in the designed architecture drops considerably since we are reusing layers instead of inserting additional new ones, (ii) the Memory Access Cost (MAC) can be reduced since the computing device can load the reused layer's parameters only once, (iii) convolutional filters get gradient update form all reuse operations, and (iv) the number of nonlinearities increases as LRU increases.

The LruNet architecture is constructed following the guidelines of recent resource efficient CNN architectures \cite{iandola2016squeezenet, rastegari2016xnor, wu2016quantized, howard2017mobilenets, sandler2018mobilenetv2, Zhang2018ShuffleNetAE, ma2018shufflenet, freeman2018effnet}. These architectures are built mostly using group convolutions and depthwise separable convolutions. Group convolutions are first introduced in AlexNet \cite{krizhevsky2012imagenet} and effectively used in ResNeXt \cite{xie2017aggregated}. Depthwise separable convolutions are proposed in Xception \cite{chollet2017xception} and they are the main building blocks for recent lightweight architectures.

Some other architectures are constructed facilitating better gradient update mechanism to get improved performance \cite{feich2018slowfast, huang2017densely, ma2018shufflenet}. Those works concentrate on how to feed the output of a layer as input to the next layers as efficiently as possible. In contrast, we reuse the layers multiple times, and as we increase the number-of-reuse (\textit{N}), convolutional filters also get more gradient updates.

\section{Approach}

\begin{figure}[t!]
	\centering
	\includegraphics[width=0.22\textwidth]{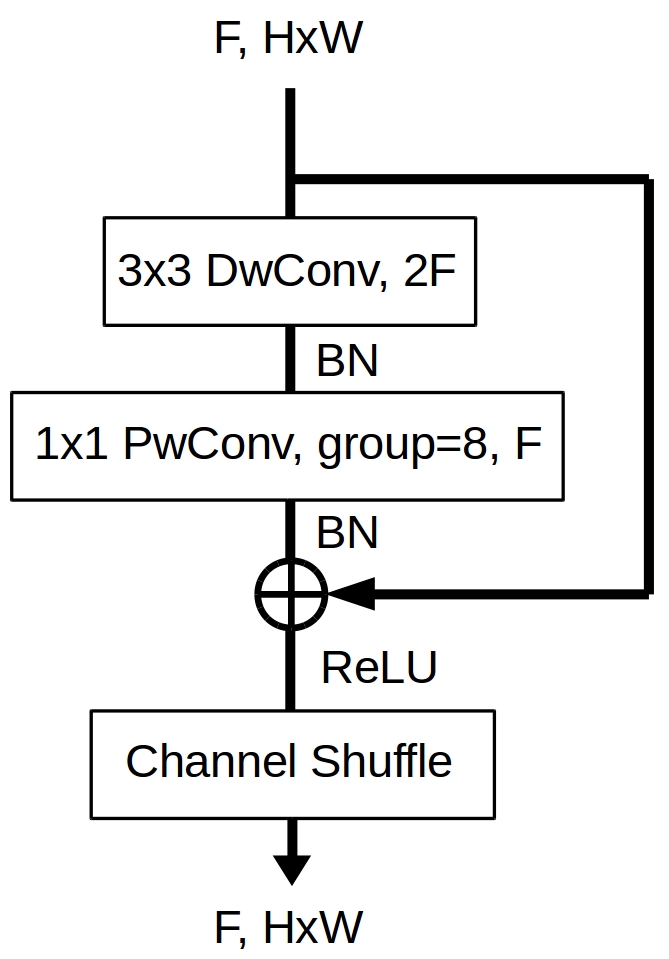}
	\caption{LRU block architecture. We represent DwConv and PwConw as depthwise and pointwise convolutions, respectively. \textit{F} is the number of feature maps of the convolutional layers and \textit{group} is the number of groups for group convolutions. \textit{BN} and \textit{ReLU} refer to Batch Normalization and Rectified Linear Unit nonlinearity, respectively. $\bigoplus$ represents summation for the shortcut connection.}
	\label{fig:block}
\end{figure}

\subsection{Layer Reuse}

Layer Reuse, referred as LRU, is the concept of using a convolutional layer or block multiple times at multiple places of a CNN architecture. However, the idea of parameter reuse in CNNs should not be limited at layer level. Convolutional layers can be split into smaller chunks up to filters, and reuse can be applied on these chunks throughout the whole network. In this case, it should be addressed as filter reuse (FRU) or kernel reuse (KRU). We believe parameter reuse concept will open a new era for deep learning practitioners for designing new CNN architectures.   

LRU block used in the LruNet network architecture is depicted in Fig. \ref{fig:block} where we used group convolutions and depthwise separable convolutions in order to keep the number of network parameters as small as possible. We first use depthwise convolutions increasing the channel number from F to 2F. Then, we applied pointwise group convolutions with 8 groups. We have tried different group numbers here, but 8 groups experimentally proved to be the best in terms of accuracy and number of parameters. We have applied Batch Normalization (BN) \cite{ioffe2015batch} after these two layers, and ReLU after the shortcut connection. Finally, channel shuffle is applied at the end of the LRU block in order to feed different channels to different filters at each reuse. Channel shuffle is implemented as in \cite{Zhang2018ShuffleNetAE}, with little modification. In the original implementation, the first and the last channel of the feature volume always remain same. So, we switched the first half of the input volume with the second half before applying shuffling. We have not applied channel shuffle at the very last reuse since there is no need.

\subsection{Network Architecture}

\begin{table}[t!]
	\centering
	\begin{tabular}{lccl}
		\specialrule{.15em}{.0em}{.1em}
		\textbf{Layer / Stride}    & \begin{tabular}[c]{@{}c@{}}\textbf{Kernel}\\\textbf{Size}\end{tabular}  & \begin{tabular}[c]{@{}c@{}}\textbf{Layer}\\\textbf{Reuse}\end{tabular} & \textbf{Output size}\\ 
		\specialrule{.15em}{.1em}{.1em}
		Conv/s2      & 3x3       &        & 64x16x16     \\
		\specialrule{.1em}{.1em}{.1em}
		LRU Block, F:64 &           & N      & 64x16x16       \\
		MaxPool/s2      & 3x3       &        & 64x8x8       \\
		Concatenate     &           &        & 128x8x8      \\
		\specialrule{.1em}{.1em}{.1em}
		LRU Block, F:128&           & N      & 128x8x8       \\
		MaxPool/s2      & 3x3       &        & 128x4x4       \\
		Concatenate     &           &        & 256x4x4      \\
		\specialrule{.1em}{.1em}{.1em}
		LRU Block, F:256&           & N      & 256x4x4       \\
		Concatenate     &           &        & 512x4x4      \\
		\specialrule{.1em}{.1em}{.1em}
		LRU Block, F:512&           & N      & 512x4x4       \\
		MaxPool/s2      & 3x3       &        & 512x2x2       \\
		\specialrule{.1em}{.1em}{.1em}
		PwConv/s1, group=8      & 1x1       &        & 256x2x2       \\
		PwConv/s1          & 1x1       &        & \textit{NumCls}x2x2\\
		AvgPool/s1      & 2x2       &        & \textit{NumCls} \\
		\specialrule{.15em}{.1em}{.0em}
	\end{tabular}
	\caption{LruNet architecture. N is the number of times LRU block is reused. PwConw refers to pointwise convolutions. Concatenation of output feature map with itself is applied at the end of each LRU Block for channel expansion, except for the last one.}
	\label{tab:lru_arch}
\end{table}

The complete LruNet network architecture is given in Table \ref{fig:lru_arch}. We have built different LruNet architectures varying the number-of-reuse \textit{N}. Similar to \cite{howard2017mobilenets}, width multiplier $\alpha$ can also be applied to scale the number of filters. The network architecture in Table \ref{fig:lru_arch} is denoted as LruNet-1x. We use \textit{N}-LruNet-$\alpha$x to denote network architecture with \textit{N} layer reuse and the number of filters in LruNet-1x are scaled by $\alpha$.

Although we reuse the convolutional layers repeatedly, we need to use a new BN layer for every reuse since the output feature volume has a different data distribution. Therefore, the number of parameters increases slightly for increased LRU due to the newly introduced BN layers. It must be also noted that the complexity and inference time of the network linearly depends on the number-of-reuse \textit{N}.

\subsection{Training Details}

All the models are trained from scratch. We use stochastic gradient descent (SGD) with mini-batch of 256, and applied categorical cross-entropy loss. For the momentum and weight decay, 0.9 and 5x10$^{-4}$ are used, respectively.

For regularization, several techniques are used to reduce overfitting. Firstly, weight decay ($\gamma$=5x10$^{-4}$) is applied on all the parameters of the network. Secondly, we used dropout before the last pointwise convolution with dropout ratio of 0.5. Lastly, we applied several data augmentation techniques: \textit{(a)} Random cropping (padding=4), \textit{(b)} random spatial rotation ($\pm$10), and \textit{(c)} random horizontal flipping.

We have trained the networks with learning rate of 0.1 for 200 epochs, and 50 more epochs with learning rates of 0.01 and 0.001. Our approach is implemented in PyTorch with a single Nvidia Titan Xp GPU.

\section{Experiments}

The proposed approach is evaluated for image classification task on three publicly available datasets: CIFAR-10, CIFAR-100 and Fashion-MNIST datasets. Each experiment is repeated 5 times in order to obtain more robust results due to the random initialization of the network parameters.

\subsection{Results Using CIFAR-10 Dataset}

\begin{table}[t!]
	\centering
	\begin{tabular}{lccc}
		\specialrule{.15em}{.0em}{.1em}
		\textbf{Model}                  & \textbf{Params}  & \textbf{MFLOPs}      & \textbf{Acc.(\%)} \\ 
		\specialrule{.15em}{.1em}{.1em}
		1-LruNet-1x	    & 131k    & 3.47 	    & 84.20            \\
		2-LruNet-1x     & 137k    & 6.30        & 84.95            \\
		4-LruNet-1x     & 149k    & 11.97       & 86.87            \\
		6-LruNet-1x     & 160k    & 17.63       & 87.91            \\
		8-LruNet-1x     & 172k    & 23.29       & 88.45            \\
		10-LruNet-1x    & 183k    & 28.95       & 88.66            \\
		12-LruNet-1x    & 195k    & 34.61       & 88.73            \\
		14-LruNet-1x    & 206k    & 40.27       & \textbf{89.34}   \\
		16-LruNet-1x    & 218k    & 45.93       & 88.45            \\
		\specialrule{.15em}{.1em}{.1em}
	\end{tabular}
	\caption{Results for different Layer Reuse (LRU) on the validation set of CIFAR-10. All networks contain the same number of convolution parameters, which is 125k.}
	\label{tab:lru_cifar10}
\end{table}

\begin{table}[t!]
	\centering
	\begin{tabular}{lc}
		\specialrule{.15em}{.0em}{.1em}
		\textbf{Model}                         & \textbf{Acc.(\%)}  \\ 
		\specialrule{.15em}{.1em}{.1em}
		8-LruNet-1x (without shuffling)              & 86.53              \\
		8-LruNet-1x (with shuffling)                 & \textbf{88.45}     \\
		\specialrule{.1em}{.3em}{.3em}
		14-LruNet-1x (without shuffling)              & 86.74            \\
		14-LruNet-1x (with shuffling)                 & \textbf{89.34}     \\
		\specialrule{.15em}{.1em}{.0em}
	\end{tabular}
	\caption{Results for Layer Reuse (LRU) with/without shuffle on the validation set of CIFAR-10.}
	\label{tab:lru_shuffle}
\end{table}

\begin{table}[b!]
	\centering
	\begin{tabular}{lcc}
		\specialrule{.15em}{.0em}{.1em}
		\textbf{Model}           & \textbf{Params}              & \textbf{Acc.(\%)}  \\ 
		\specialrule{.15em}{.1em}{.1em}
		67-depth Network ($\sim$ 8-LruNet-1x)       & \textbf{172k}       & 88.45              \\
		67-depth Network (with new layers)    & 902k       & \textbf{90.27}     \\
		\specialrule{.1em}{.3em}{.3em}
		115-depth Network ($\sim$ 14-LruNet-1x)      & \textbf{206k}       & 89.34            \\
		115-depth Network (with new layers)   & 1562k      & \textbf{90.93}     \\
		\specialrule{.15em}{.1em}{.0em}
	\end{tabular}
	\caption{Comparison of LruNet with the networks having same depth containing all newly introduced layers.}
	\label{tab:same_depth}
\end{table}

The CIFAR-10 \cite{krizhevsky2009learning} is a fundamental dataset in computer vision containing 50k training and 10k testing images in 10 classes with image resolution of 32x32. Initially we investigate the effect of LRU on the performance. Results in Table \ref{tab:lru_cifar10} show that acquired accuracy increases as we increase the applied LRU until 14-LRU. Afterwards, the performance does not improve with further reusing. As it has been mentioned earlier, the computational complexity (floating point operations - FLOPs) depends on the number-of-reuse \textit{N} linearly. 14-LruNet-1x achieves 5.14\% better classification accuracy than 1-LruNet-1x.

Secondly, we investigate the effect of channel shuffle. Results in Table \ref{tab:lru_shuffle} show that channel shuffle plays an important role in layer reuse. For both 8-LruNet-1x and 14-LruNet-1x cases, networks with channel shuffle perform better compared to the networks without channel shuffle. It is also interesting to note that even without channel shuffle, networks perform better than 1-LruNet-1x.

Lastly, Table \ref{tab:same_depth} compares the LruNet with the networks having same depth containing all newly introduced layers. Although networks with newly introduced layers achieves comparatively better results, networks with LRU have 5.24 and 7.58 times less parameters for 67-depth and 115-depth networks, respectively.

\begin{table}[t!]
	\centering
	\begin{tabular}{lccc}
		\specialrule{.15em}{.0em}{.1em}
		\textbf{Model}    & \textbf{Params}  & \textbf{MFLOPs}  & \textbf{Acc.(\%)} \\ 
		\specialrule{.15em}{.1em}{.1em}
		1-LruNet-2x	     & 514k       & 10.44     & 63.02           \\
		2-LruNet-2x      & 525k       & 19.77     & 64.77           \\
		4-LruNet-2x      & 549k       & 38.44     & 65.50           \\
		6-LruNet-2x      & 572k       & 57.10     & 66.02           \\
		8-LruNet-2x      & 595k       & 75.76     & 67.57           \\
		10-LruNet-2x     & 618k       & 94.42     & 67.98           \\
		12-LruNet-2x     & 641k       & 113.08    & 68.30           \\
		14-LruNet-2x     & 664k       & 131.74    & \textbf{68.87}           \\
		16-LruNet-2x     & 687k       & 150.40    & 68.02           \\
		\specialrule{.15em}{.1em}{.0em}
	\end{tabular}
	\caption{Results for different Layer Reuse (LRU) on the validation set of CIFAR-100. All networks contain the same number of convolution parameters, which is 501k.}
	\label{tab:lru_cifar100}
\end{table}

\begin{table}[t!]
	\centering
	\begin{tabular}{lccc}
		\specialrule{.15em}{.0em}{.1em}
		\textbf{Model}                         & \textbf{Params}  & \textbf{MFLOPs}  & \textbf{Acc.(\%)} \\ 
		\specialrule{.15em}{.1em}{.1em}
		1-LruNet-1x	    & 130k	     & 2.85     & 91.17           \\
		2-LruNet-1x     & 136k       & 5.40     & 92.28           \\
		4-LruNet-1x     & 148k       & 10.50    & 92.69           \\
		6-LruNet-1x     & 159k       & 15.61    & 93.13           \\
		8-LruNet-1x     & 171k       & 20.71    & 93.27           \\
		10-LruNet-1x    & 182k       & 25.81    & \textbf{93.46}  \\
		12-LruNet-1x    & 194k       & 30.92    & 93.34           \\
		\specialrule{.15em}{.1em}{.0em}
	\end{tabular}
	\caption{Results for different Layer Reuse (LRU) on the validation set of Fashion-MNIST. All networks contain the same number of convolution parameters, which is 124k.}
	\label{tab:lru_fashionmnist}
\end{table}

\subsection{Results Using CIFAR-100 Dataset}

CIFAR-100 \cite{krizhevsky2009learning} is very similar to the CIFAR-10, except it has 100 classes containing 600 images each. Since it is a more challenging task, we have increased the dropout rate to 0.7 and used width multiplier $\alpha$ of 2. We again analyzed the effect of LRU on the performance, and 14-LruNet-2x achieves 5.14\% better classification accuracy than 1-LruNet-2x.

\subsection{Results Using Fashion-MNIST Dataset}

Fashion-MNIST \cite{xiao2017fashion} is very similar to MNIST \cite{lecun1998gradient}, but contains images of various articles of clothing and accessories. There are 50k training and 10k testing images in grayscale for 10 classes with image resolution of 28x28. 10-LruNet-1x achieves 2.29\% better classification accuracy than 1-LruNet-1x. This performance improvement is relatively small compared to the performance improvements using CIFAR-10 and CIFAR-100 This might be due to the relative simpleness of the Fashion-MNIST dataset compared to the CIFAR-10 and CIFAR-100. 1-LruNet-1x already achieves 91.17\% classification accuracy on Fashion-MNIST.

Table \ref{tab:sota} shows the comparison of LruNet with the state-of-the-art results. For ShuffleNetv2, channel numbers are adjusted accordingly for width multipliers of 0.25 and 0.75. Results in Table \ref{tab:sota} show that LruNet achieves comparatively similar results, although it has much smaller number of convolutional parameters. 

\begin{table}[]
    \centering
    \begin{tabular}{clccc}
    \specialrule{.15em}{.0em}{.1em}
        & \textbf{Model}  & \begin{tabular}[c]{@{}c@{}}\textbf{Total}\\\textbf{Params}\end{tabular} &\begin{tabular}[c]{@{}c@{}}\textbf{Conv.}\\\textbf{Params}\end{tabular} & \textbf{Acc.(\%)} \\     
        \specialrule{.15em}{.1em}{.1em}
        \multicolumn{1}{c|}{\multirow{5}{*}{\rotatebox[origin=c]{90}{\textbf{CIFAR10}}}} & ShuffleNet-0.5x(g=3) & 229k   & 217k    & \textbf{90.32}    \\  
        \multicolumn{1}{c|}{}                      & ShuffleNetV2-0.25x    & 211k     & 205k       & 89.31    \\ 
        \multicolumn{1}{c|}{}                      & MobileNet-0.25x       & 216k     & 210k       & 84.72    \\  
        \multicolumn{1}{c|}{}                      & MobileNetV2-0.25x     & 249k     & 239k       & 89.57    \\  
        \multicolumn{1}{c|}{}                      & 14-LruNet-1x          & 206k     & \textbf{125k}       & 89.34    \\ 
        \specialrule{.15em}{.1em}{.1em}
        \multicolumn{1}{c|}{\multirow{5}{*}{\rotatebox[origin=c]{90}{\textbf{CIFAR100}}}} & ShuffleNet-0.75x(g=3) & 560k  & 542k    & \textbf{69.97}    \\  
        \multicolumn{1}{c|}{}                      & ShuffleNetV2-0.75x    & 568k     & 559k       & 69.15    \\ 
        \multicolumn{1}{c|}{}                      & MobileNet-0.4x        & 567k     & 558k       & 60.54    \\ 
        \multicolumn{1}{c|}{}                      & MobileNetV2-0.4x      & 603k     & 588k       & 69.95    \\  
        \multicolumn{1}{c|}{}                      & 14-LruNet-2x          & 664k     & \textbf{501k}       & 68.87    \\ 
        \specialrule{.15em}{.1em}{.1em}
        \multicolumn{1}{c|}{\multirow{5}{*}{\rotatebox[origin=c]{90}{\textbf{Fash.-MNIST}}}} & ShuffleNet-0.5x(g=3) & 228k & 216k   & \textbf{94.11}    \\  
        \multicolumn{1}{c|}{}                      & ShuffleNetV2-0.25x    & 211k     & 205k       & 93.42    \\ 
        \multicolumn{1}{c|}{}                      & MobileNet-0.25x       & 216k     & 210k       & 90.63    \\  
        \multicolumn{1}{c|}{}                      & MobileNetV2-0.25x     & 249k     & 239k       & 93.43    \\  
        \multicolumn{1}{c|}{}                      & 10-LruNet-1x          & 182k     & \textbf{124k}       & 93.46    \\ 
        \specialrule{.15em}{.1em}{.0em}
    \end{tabular}
    \caption{Comparison of LruNet with state-of-the-art results.}
	\label{tab:sota}
\end{table}
\section{Conclusion}

This paper proposes a parameter reuse strategy, Layer Reuse (LRU), where convolutional layers of a CNN architecture (LruNet) are used repeatedly. We evaluated the LRU on several publicly available datasets and achieved improved classification performance. LRU especially boosts the CNNs with small number of parameters which is of utmost importance for embedded applications. We believe this work will open up a new research direction for deep learning practitioners for designing novel CNN architectures.

As a future work, we would like to analyze different parameter reuse strategies. It must be noted that parameter reuse is not only restricted to layer level, but we can also apply Filter Reuse (FRU) (or Kernel Reuse (KRU)) for efficient CNN architecture designs.

\section*{Acknowledgements}
We gratefully acknowledge the support of NVIDIA Corporation with the donation of the Titan Xp GPU used for this research.

\bibliographystyle{IEEEbib.bst}
\bibliography{REFS}

\end{document}